\documentclass{article}

\usepackage[preprint]{neurips_2024}

\usepackage[utf8]{inputenc} 
\usepackage[T1]{fontenc}    
\usepackage{hyperref}       
\usepackage{url}            
\usepackage{booktabs}       
\usepackage{amsfonts}       
\usepackage{nicefrac}       
\usepackage{microtype}      
\usepackage{xcolor}         
\usepackage{pdfpages}
\usepackage{graphicx}
\usepackage{subcaption}
\usepackage{algorithm}
\usepackage{algpseudocode}
\usepackage{placeins} 
\usepackage{natbib}
\usepackage{amsmath} 
\usepackage{makecell}
\usepackage{array} 
\usepackage{hyperref}
\usepackage{authblk}
\title{Correlation-Aware Select and Merge Attention for Efficient Fine-Tuning and Context Length Extension}

\author{
  Ning Wang$^{1,2,3}$\thanks{First Author: \texttt{wangning2023@ia.ac.cn}} \quad Zekun Li$^{1,2,3}$ \quad Tongxin Bai$^{3}$\thanks{Corresponding Author: \texttt{txbai@baai.ac.cn}} \quad Guoqi Li$^{1,2}$\thanks{Corresponding Author: \texttt{guoqi.li@ia.ac.cn}} \\
  $^1$Institute of Automation, Chinese Academy of Sciences, Beijing, China \\
  $^2$University of Chinese Academy of Sciences, Beijing, China \\
  $^3$Beijing Academy of Artificial Intelligence, Beijing, China
}
\begin{document}

\maketitle

\begin{abstract}
    Modeling long sequences is crucial for various large-scale models; however, extending existing architectures to handle longer sequences presents significant technical and resource challenges. In this paper, we propose an efficient and flexible attention architecture that enables the extension of context lengths in large language models with reduced computational resources and fine-tuning time compared to other excellent methods. Specifically, we introduce correlation-aware selection and merging mechanisms to facilitate efficient sparse attention. In addition, we also propose a novel data augmentation technique involving positional encodings to enhance generalization to unseen positions. The results are as follows: First, using a single A100 40GB GPU, we achieve fine-tuning on Llama2-7B with a sequence length of 32K, which is more efficient than other methods that rely on subsets for regression. Second, we present a comprehensive method for extending context lengths across the pre-training, fine-tuning, and inference phases. During pre-training, our attention mechanism partially breaks translation invariance during token selection, so we apply positional encodings only to the selected tokens. This approach achieves relatively high performance and significant extrapolation capabilities. For fine-tuning, we introduce Cyclic, Randomly Truncated, and Dynamically Growing NTK Positional Embedding (CRD NTK). This design allows fine-tuning with a sequence length of only 16K, enabling models such as Llama2-7B and Mistral-7B to perform inference with context lengths of up to 1M or even arbitrary lengths. Our method achieves 100\% accuracy on the passkey task with a context length of 4M and maintains stable perplexity at a 1M context length. This represents at least a 64-fold reduction in resource requirements compared to traditional full-attention mechanisms, while still achieving competitive performance. Our code are available at \href{https://github.com/fisalt/MS-Attention}{GitHub Repository}
\end{abstract}

\section{Introduction}

In various natural language processing (NLP) tasks, such as document-level sentiment analysis\cite{behdenna2018document}, long document summarization\cite{10.1145/3545176}, and code generation\cite{rozière2024code}, the ability to model long-sequence dependencies effectively is crucial. This capability allows for capturing complex relationships over sequences spanning hundreds or thousands of tokens, which is essential for tasks where contextual information is dispersed. For example, accurately summarizing a lengthy document or generating coherent code relies on understanding dependencies that go beyond adjacent tokens. Consequently, extending the context window enables LLMs to perform tasks that shorter context windows cannot handle and potentially enhances performance across a variety of NLP tasks.

However, extending the context window presents numerous challenges. Longer sequences demand significantly more memory and computational resources, leading to slower training and inference times and higher resource consumption. Moreover, capturing long-range dependencies and using large numbers of tokens in autoregressive models can result in slower convergence, potentially due to underfitting caused by the large number of tokens involved in the autoregressive process.

To address these challenges, current research often adopts a strategy of pretraining on short sequences\cite{touvron2023llamaa}\cite{jin2023growlength} followed by efficient fine-tuning using positional interpolation or positional extrapolation for long sequence extension. Recent works have achieved promising results, such as LongLora\cite{chen2024longlora}, which combines sparse attention with improved LoRA\cite{hu2021lora} (Low-Rank Adaptation) to extend Llama2-7B\cite{touvron2023llamab} to 100K tokens using 8$\times$A100 80GB GPUs. However, our analysis indicates that LongLora’s sparse attention does not fully exploit sparsity, and its efficient fine-tuning can be further optimized to use fewer parameters, thereby requiring less memory and computational resources for similar extensions.

In this paper, we propose a novel approach that leverages correlation-based selection and merging mechanisms to achieve more efficient sparse attention, along with fine-tuning strategies that optimize parameter utilization. Our attention mechanism achieves a degree of controlled fitting by selecting variable token quantities, adjusting receptive fields, and configuring varying levels of compression. This flexibility allows our method to be effectively applied to fine-tuning across different datasets and domains, such as supervised fine-tuning (SFT). To exploit this capability further, we design a cyclic, randomly sampled, and dynamically expanding NTK (CRD NTK) positional encoding during fine-tuning to generalize positional information. This design facilitates significant extrapolation capabilities and theoretically supports arbitrary extensions of the context window. Moreover, in addition to length extension via fine-tuning, we present theoretical and empirical evidence that our method can disrupt positional invariance to a certain degree, enabling efficient automatic extrapolation during pre-training. Our method also integrates effectively with approaches like InfLLM\cite{xiao2024infllmtrainingfreelongcontextextrapolation}, allowing for long-sequence inference without requiring additional fine-tuning. As such, we offer a comprehensive framework for extending context windows across the pre-training, fine-tuning, and inference stages.

\textbf{Efficiency:} Our method utilizes a single A100 GPU in conjunction with DeepSpeed ZeRO-2\cite{10.1145/3394486.3406703} to fine-tune Llama2-7B for a context length of 32K, demonstrating that other methods, such as LongLora\cite{chen2024longlora}, encounter memory overflow issues. We provide algorithmic complexity analysis and memory consumption metrics during a model pre-training process. Furthermore, our approach operates orthogonally to methods like RingAttention\cite{liu2023ringattentionblockwisetransformers}.

\textbf{Context Length Extension of Positional Encoding:} The paper points out that Full Attention, MS Attention, or other attention mechanisms can achieve high-ratio positional extrapolation through fine-tuning using either high-scale factor positional interpolation (PI) or NTK positional encoding. As a result, the length extension proposed in this paper is no longer limited by setting the PI or NTK scale factor parameters based on the ratio between pre- and post-extension lengths. Instead, we fine-tune using large scale factor parameters of our own choosing. Furthermore, based on theoretical analysis, we designed a CRD NTK positional encoding that better matches our attention mechanism (we believe a similar property applies to PI positional encoding as well). Thus, by fine-tuning with a context length of 16K, our method achieves context window extensions exceeding 1M for both Llama2-7B and Mistral-7B, yielding 100\% accuracy on the passkey task at a context length of 4M and maintaining stable perplexity in text testing at 1M length. While further testing for longer lengths was limited due to resource constraints, theoretical and experimental analyses indicate that our method can support arbitrary context window extensions for both relative and partial absolute positional encodings. Our method intrinsically disrupts position invariance, allowing for substantial context window length extrapolation through the designed positional encodings.

\textbf{Flexible Fitting Degree:} Our approach implements different quantities of selected and merged tokens, as well as varying degrees of compression and receptive field ranges, to achieve diverse convergence behaviors on the Llama2-7B model, resulting in different perplexity levels. 

\begin{figure}[ht]
\vspace{-0.5cm}
  \centering
  \label{selection_merge}
  \includegraphics[width=\linewidth]{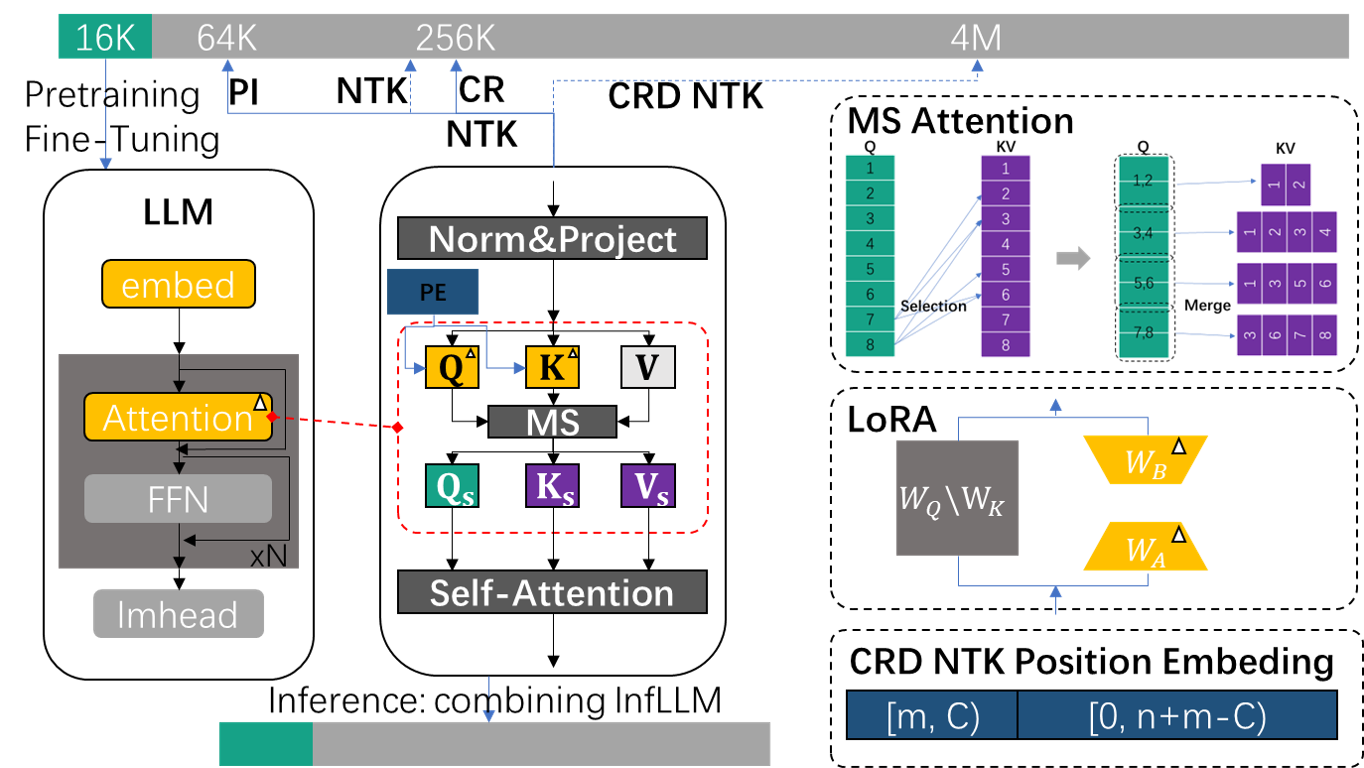}
  \caption{Overview of Efficient Fine-Tuning and Context Length Extension. Efficient fine-tuning is achieved using the Merge and Select Attention Mechanism (MS Attention) combined with LoRA. For length extension during fine-tuning, we build upon this efficient approach by employing Cyclic, Randomly Truncated, and Dynamically Growing NTK Positional Embedding (CRD NTK) for high-order extrapolation. The pre-training method capable of high-order length extrapolation utilizes MS Attention along with CRD NTK. Finally, for direct inference without fine-tuning, we adopt the methodology from InfLLM.}
\vspace{-0.5cm}
\end{figure}

In summary, we achieve efficient attention through a combination of selection and merging mechanisms. This algorithm, when combined with specially designed positional encoding methods or other existing techniques, enables substantial context length extensions across pre-training, fine-tuning, and inference phases. Additionally, we demonstrate that by tuning parameters such as token selection, merging, compression, and receptive fields, our method allows for  a certain degree of controlled fitting. This controllability is a key factor enabling large models to generalize positional information effectively during fine-tuning, one of the essential capabilities of our approach.


\section{Related Work}

\textbf{Efficient Attention Mechanisms}
To fully exploit the inherent sparsity and positional relationships between tokens, a significant body of research has focused on developing efficient attention mechanisms. These mechanisms reduce the computational complexity of attention operations by focusing on a subset of tokens at each step, thus processing long sequences more efficiently or reducing resource consumption. Existing methods first preserve local features and then use various strategies to attend to more distant tokens. For instance, BigBird\cite{NEURIPS2020_c8512d14} and Performer\cite{choromanski2022rethinking} use random patterns, Longformer\cite{beltagy2020longformer} and DETR\cite{zhu2021deformable} use fixed patterns, while Biformer\cite{Zhu_2023_CVPR} and Routing Transformer\cite{roy-etal-2021-efficient} utilize relevance routing mechanisms. Our proposed relevance selection and merging mechanism adapts flexibly to various scenarios and is compatible with FlashAttention2\cite{dao2022flashattention}, achieving more efficient and general sparse attention.

\textbf{Positional Encoding}
Another research direction aimed at extending sequence length in LLMs focuses on positional encoding techniques, such as positional interpolation and extrapolation. Most pretrained models are trained on fixed-length sequences with fixed positional encodings, leading to performance degradation when extended to unknown positions. Therefore, numerous studies have analyzed the impact of positional encodings and modified them through interpolation or extrapolation to extend to longer sequences. For example, Position Interpolation\cite{chen2023extending}, NTK-aware\cite{Ntkaware}, Yarn\cite{peng2023yarn}, and LongRopE\cite{ding2024longrope} mitigate the effects of pretrained positional encodings by using interpolation with different scales based on frequency importance, effectively extending sequence modeling lengths.

\textbf{Efficient Fine-Tuning}
Efficient fine-tuning of LLMs has become a critical research direction, especially for handling long sequences. Techniques like Input-tuning\cite{an2022inputtuning} and LoRA\cite{hu2021lora} have shown significant promise in this area. Building on the LongLora\cite{chen2024longlora} method, we further optimize by using fewer parameters for fine-tuning, aiming to reduce the computational and memory overheads associated with fine-tuning large models on extended sequences while maintaining their performance on downstream tasks.


\section{Preliminary}

\subsection{Transformer}

The Llama2 model used in this paper is based on the Transformer architecture, which consists of the core modules self-attention and MLP. The computation process of self-attention\cite{vaswani2017attention} is as follows:
\[ O = softmax(QK^T)V \]
where \( Q \), \( K \), and \( V \) are obtained from \( X \) using embedding weights \( W_q \), \( W_k \), and \( W_v \) respectively. The final output \( O \) is then passed through \( W_o \) to obtain the final output of the attention. Subsequently, the entire Transformer operation is completed through the MLP. 

\subsection{LoRA}

Low-Rank Adaptation (LoRA)\cite{hu2021lora} is an efficient model fine-tuning method designed to reduce the computational resources and storage requirements when fine-tuning large pretrained models.

The core idea of LoRA is to decompose the weight matrix \( W \) of the pretrained model into two low-rank matrices, \( A \) and \( B \). This decomposition is represented as \( W = W_0 + \Delta W \), where \( W_0 \) is the original weight matrix and \( \Delta W = A B \) is the adjustment matrix obtained through low-rank decomposition. The matrices \( A \) and \( B \) have a rank of \( r \), which is typically much smaller than the dimensions of \( W \). During fine-tuning, only the parameters of matrices \( A \) and \( B \) need to be adjusted, while the original weight matrix \( W_0 \) remains unchanged.

The LoRA experiments on various large language models (such as GPT-3\cite{brown2020language} and BERT\cite{devlin2018bert}) have shown that the method can significantly reduce the computational resources and storage requirements while maintaining or even improving the model's performance.

\subsection{Positional Encoding}\label{Positional Encoding}

Positional encoding can be divided into relative and absolute positional encodings. The most widely used method for relative positional encoding is the Rotary Positional Encoding (RoPE). The encoding formula is given as follows, where \(\theta_i = \text{base}^{-\frac{2i}{d}}\):
\[
PC_{m} = 
\begin{bmatrix}
\cos(m\theta_0) & \cos(m\theta_0) & \cos(m\theta_1) & \cos(m\theta_1) & \dots & \cos(m\theta_{d/2-1}) & \cos(m\theta_{d/2-1})
\end{bmatrix}
\]
\[
PS_{m} = 
\begin{bmatrix}
\sin(m\theta_0) & -\sin(m\theta_0) & \sin(m\theta_1) & -\sin(m\theta_1) & \dots & \sin(m\theta_{d/2-1}) & -\sin(m\theta_{d/2-1})
\end{bmatrix}
\]

The output for the encoding of the \(m\)-th token is given by:

\[
X_m' = 
\begin{bmatrix}
x_0 & x_1 & x_2 & x_3 & \dots & x_{d-1} & x_{d}
\end{bmatrix} \odot PC_{m} + 
\begin{bmatrix}
x_1 & x_0 & x_3 & x_2 & \dots & x_{d} & x_{d-1}
\end{bmatrix} \odot PS_{m}
\]
\textbf{Positional Interpolation}\cite{chen2023extending} is performed by scaling the above position, where the index of the \(m\)-th position is transformed as follows:
\[
Pf_{m} = 
\begin{bmatrix}
f(\frac{m}{\lambda}\theta_0) & f(\frac{m}{\lambda}\theta_0) & f(\frac{m}{\lambda}\theta_1) & f(\frac{m}{\lambda}\theta_1) & \dots & f(\frac{m}{\lambda}\theta_{d/2-1}) & f(\frac{m}{\lambda}\theta_{d/2-1})
\end{bmatrix}
\]
\textbf{NTK Positional Encoding}\cite{Ntkaware}: For NTK positional encoding, the base is rescaled as:
\[
\text{base}' = \text{base} \times \left(\text{scale}^{\frac{d}{d-2}}\right), \theta_i = \text{base}'^{-\frac{2i}{d}}
\]
\textbf{Other Positional Encodings}: Other positional encoding methods extend from the two encoding strategies described above. For example, YaRN\cite{peng2023yarn} posits that high-frequency information is more important, so it retains the high-frequency portion of the positional encoding unchanged, while applying interpolation to the low-frequency components. LongRoPE\cite{ding2024longrope} addresses the issue of attention "sink" by preserving the positional encoding up to a specific position \(\hat{n}\), and performing interpolation for positions beyond \(\hat{n}\). The interpolation ratio is determined via a search algorithm.

\section{Methods}

\subsection{MS: Merge selection}\label{MS}
The following is a pseudo-code and the theory in appendix \ref{Efficient Attention Methods}:

\begin{algorithm}
\caption{Selection and Merging Process}
\begin{algorithmic}[1]
\State \textbf{Input:} Q,K,V tensors of shape $(b, h, n, d)$, segment size $s$, top$k$, top$n$, $merges$
\State \textbf{Output:} Final indices $kvm$

\Procedure{Selection}{}
\Comment{Selection}
    \State Partition Q tensors into regions: $(b, h, n, d) \rightarrow (b, h, n_{sq}, s_q, d)$
    \State Partition KV tensors into regions: $(b, h, n, d) \rightarrow (b, h, n_{sk}, s_k, d)$
    \State Represent each region with a semantic or averaged compressed token: $Q_s'\in(b, h, n_{sq}, d), K_s'\in(b, h, n_{sk}, d)$
    \State Compute relevance between $Q_s'$ and $K_s'$ using dot product or other similarity metrics
    \State Apply mask to prevent information leakage
    \State Obtain indices of top $k$ most relevant $K_s$ regions: $selectindx \in (b, h, n_{sq}, \text{topk})$
\EndProcedure

\Procedure{Merging}{}
\Comment{Merging}
    \State Merge $Q_s$ regions: $(b, h, ns, s, d) \rightarrow (b, h, n_{ms}, m \cdot s, d), m=merges$
    \State Split, permute, and merge $selectindx$: $mselectindx \in (b, h, n_{ms}, m, \text{topk}) \rightarrow (b, h, n_{ms}, \text{topk}, m) \rightarrow (b, h, n_{ms}, \text{topk} \cdot m)$
    \State Perform unique operation while preserving relevance order
    \State Select top $n$ indices: $qmselectindx$ representing final set of relevant $K_s$ regions
    \State Select relevant $KV_s$ regions by $qmselectindx$: $kvm\in(b, h, n_{sq}, topn, d)$
\EndProcedure

\end{algorithmic}
\end{algorithm}

We propose a method for implementing more general and efficient sparse attention through correlation selection and merging mechanisms, as shown in Figure \ref{selection_merge}. This method consists of two main steps: selection of relevant regions and merging of these regions.
The proposed method consists of two main steps: selection and merging.

\begin{figure}[t]
  \centering
  \includegraphics[width=0.75\linewidth]{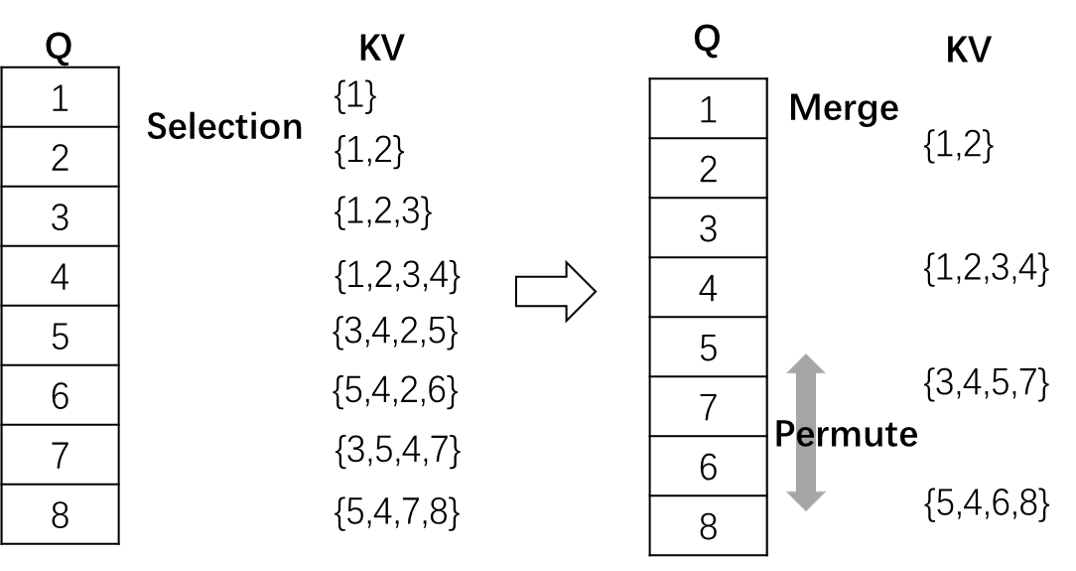}
  \caption{\textbf{Overview of Merging and Selection Attention Mechanism (MS Attention).} The MS Attention mechanism involves two main steps. In the first step, the QKV tensors are split into regions, and a single token is used to represent each region. Subsequently, the regional representatives are used to compute dot products or other similarity measures to select the most relevant KV regions for each Q region. For example, Q regions 5, 6, 7 and 8 select KV regions \(\{3, 4, 2, 5\}\), \(\{5, 4, 2, 6\}\), \(\{3, 5, 4, 7\}\) and \(\{5, 4, 7, 8\}\) respectively. In the second step, for each Q region, tokens are merged with their adjacent or related tokens after permuting. The union of the selected KV regions is taken, and the top-n regions are chosen. For example, combine Q regions 6 and 8, along with their selected KV regions, resulting in \(\{5, 4, 2, 7, 6, 8\}\). To ensure tensor consistency, we select the top \(k\) regions from the merged set. If \(k = 4\), the final selection is \(\{5, 4, 6, 8\}\). The reason for retaining 6 and 8 is because by default we believe that the local region is important and must be preserved, and that the local region does not perform scoring calculations. Finally, each merged Q region performs self-attention with its selected relevant KV regions.}
  \label{selection_merge}
\end{figure}

\textbf{Step 1: Selection.} First, the \( Q \), \( K \), and \( V \) tensors with shape \((b, h, n, d)\) are segmented into regions, resulting in tensors \( Q_s \) with shape \((b, h, n_{sq}, s_q, d)\), \( K_s \), and \( V_s \) with shape \((b, h, n_{sk}, s_k, d)\), where \( s_q \)  and \( s_k \) denote the segment size. Each region is represented by a semantic token or an average compressed token, yielding \( Q_s' \) and \( K_s' \). A dot product or another similarity metric is then applied between \( Q_s' \) and \( K_s' \) to analyze the relevance between \( Q_s \) and \( K_s \) regions. To prevent information leakage, we apply a mask and control the number of selected tokens. This process results in the indices of the top-\( k \) most relevant \( K_s \) regions for each \( Q_s \) region, denoted as \( selectindx \) with shape \((b, h, n_{sq}, topk)\), where \( n_s \) is the number of \( Q_s \) regions and \(topk\) is the number of selected \( K_s \) and \( V_s \) regions.

\textbf{Step 2: Merging.} In the merging step, the indices obtained from the first step and \( Q_s \) are permuted and combined, effectively merging the selected regions. \( Q_s \) is merged according to a specified number of segments \( merges \), resulting in \( Q_{ms} \) with shape \((b, h, n_{ms}, m \cdot s, d)\). Similarly, \( selectindx \) is split, permuted, and merged into \( mselectindx \) with shape \((b, h, n_{ms}, topk, m) \rightarrow (b, h, n_{ms}, topk \cdot m) \). Due to previous topk operation, the permutation can ensure that each row is sorted by relevance, with the first \(m\) indices corresponding to the most relevant \(K_s\) region for each of the merged \(Q_s\) regions, the next \(m\) indices corresponding to the second most relevant \(K_s\) region, and so on. After obtaining the merged indices \( \text{mselectindx} \), performs unique operation while maintaining relevance order. The top-\( n \) indices, denoted as \( \text{qmselectindx} \), are selected as the final indices corresponding to the merged \( Q \) regions and their relevant \( K_s \) regions.

For pre-training and inference, we apply a permutation transformation to \( Q_s \) and indices based on the correlation in the selection step. This enables similar regions of \( Q_s \) to share more common \( KV \) regions, significantly enhancing performance. The permutation transformation is derived as follows: for the selection step indices \( selectindx \), we apply a weighted sum based on a chosen base. In our experiments, we select a base of \( b=5 \), and use the first \( n=10 \) indices as the criterion. The score \( s \) for each row of indices corresponding to a region of \( Q_s \) is calculated as:

\[
s = \sum_{i=0}^{n-1} b^{n-i} \times \frac{selectindx[i]}{max}
\]

We then sort based on the calculated scores, where the permutation of positions before and after sorting represents the desired transformation. After computing the output, we apply the inverse of this permutation to restore the original order of the \( Q_s \) regions. For fine-tuning with extended context lengths, we consider omitting the permutation step, as the main goal is to employ the selection mechanism to regress the next tokens using partial positional information.

The reason for this merging step is to ensure that each token can attend to enough \( K \) and \( V \) tokens, and using larger \( Q \) regions sharing more \( K \) and \( V \) tokens is more efficient than smaller regions with fewer \( K \) and \( V \) tokens. Another reason for using the merging strategy is that relevance selection within smaller regions tends to be more precise.

While sharing similarities with Biformer\cite{Zhu_2023_CVPR} and Routing Transformer\cite{roy-etal-2021-efficient}, our method distinguishes itself through its flexible representation of regions, the adaptability and compressibility in selecting region sizes, and the additional versatility of the merger mechanism. This allows for a more nuanced and effective approach to handling attention mechanisms in large-scale language models, enhancing their performance and efficiency across various tasks and contexts. For example, by flexibly setting the size of the \( Q \), \( K \), and \( V \) partitioned regions in these two steps, the range of  \( KV \) attended by \( Q \) tokens, the number of selected \( K \) and \( V \) tokens, the size of the merged regions and the number of retained \( K \) and \( V \) tokens, our attention mechanism can cover almost all scenarios.

As analyzed in Appendix \ref{Efficient Attention Methods}, our algorithm is able to cover most of the autoregressive methods using a subset of KV tokens, examples include Biformer, Landmark Attention, Routing Transformer and Swin Transformer and their variants, etc.

\subsection{Reduced LongLora}\label{Reduced LongLora}

We propose a method that selectively fine-tunes only the \( W_k \) and \( W_o \) weights, achieving results nearly identical to fine-tuning the entire attention mechanism. Specifically, since \( QK^T = X W_q W_k^T X^T \), updating only \( W_k \) yields \( W_q W_k^T \), effectively replicating the effect of simultaneously updating \( W_k \) and \( W_q \). This approach is particularly effective when \( W_q \) is full rank. Similarly, fine-tuning \( W_o \) follows the same rationale, as the final output is a linear mapping of \( W_v W_o \). Since only the \(Q\) and \(K\) tokens undergo positional encoding, the weights to be fine-tuned are limited to \(W_Q\) and \(W_K\). Based on the above analysis, it suffices to only fine-tune \(W_Q\) or \(W_K\). In this work, for the fine-tuning of LLaMA, we adopt a low-rank fine-tuning of \(W_Q\) and \(W_K\), while for Mistral, we apply low-rank learning of \(W_Q\) for sequence length extension.

Additionally, since query tokens can be considered well-fitted through extensive training, learning the linear mapping for their corresponding key tokens is reasonable. Furthermore, as the attention mechanism becomes heterogeneous, updating the final classification head is a direct approach. However, considering the large number of parameters, altering the initial embedding\cite{an2022inputtuning} to achieve a certain degree of equivalence is also a feasible consideration.

\subsection{Cyclic, Randomly Truncated, and Dynamically Growing NTK Method}\label{NTK Method}

For positions not encountered during pretraining, it is necessary to scale these positions within the range of positions learned during pretraining. In this work, we employ the NTK method for scaling, adjusting the positional base proportionally. During fine-tuning, after a certain number of steps, the base is dynamically altered to accommodate different scales.

To address the issue of attention sink, we apply random shifts and cyclic modular operations. This approach not only alleviates the attention sink problem but also enables data augmentation of positional information, which improves generalization capabilities. By leveraging the model’s more flexible fitting ability, we can better learn and generalize positional information. The positional encoding is as follows:
\[
Pf_{mi} = f\left(\left((m + \text{randomp}) \% \text{max}\right) \left(\text{base} \times \left(\text{scale}^{\frac{d}{d-2}}\right)\right)^{-\frac{2i}{d}}\right)
\]
The analysis of backpropagation for learning positional information is provided in the appendix.

\subsection{The strategy for incorporating positional information during pre-training}

Our algorithm leverages semantic token routing, where each semantic token encapsulates contextual information from its surrounding tokens, similar to a sliding window mechanism in convolution. Since different tokens are surrounded by distinct contexts, the resulting semantic tokens are unique, leading to different routed \(KV\) tokens and interpolated variables. As a result, we do not assign positional information during the selection process. Instead, after the selection process, RoPE positional encoding is applied to the \(K\) and \(Q\) tokens to enhance performance. 

\section{Experiment}

\subsection{Experimental Setup}

Our experiments were conducted on 2 $\times$ A100 40GB GPUs, using the Llama2-7B\cite{touvron2023llamab} model and Mistral-7B-v0.1 with the attention mechanism replaced by our MS Attention as described in Section \ref{MS}. The training approach used the efficient fine-tuning method described in Section \ref{Reduced LongLora}, with position interpolation applied. Finally, to extend to longer sequence lengths, we employed the recursive method described in Section \ref{MS recursion}.

\textbf{Fine-tuning Steps and Parameters:} The fine-tuning parameters included the linear transformations for key and output ($W_k$ and $W_o$), as well as the embedding and normalization parameters. The training approach used an autoregressive method, where the objective was to generate the next token. The loss function used was cross-entropy, and the optimizer was AdamW\cite{loshchilov2019decoupled} with parameters $\beta_1$ = 0.9, $\beta_2$ = 0.95 and $lr=1e-4$. Both the batch size and gradient accumulation were set to 1, and the model was trained for 3000 steps.

\textbf{Dataset:} The dataset used for fine-tuning was the Redpajama\cite{together2023redpajama} dataset, which is the same dataset used in LongLora. Evaluation was performed using the widely used long text dataset PG19\cite{raecompressive2019} and proof-pile, with perplexity as the evaluation metric. Additionally, the model's performance was assessed using the passkey task\cite{mohtashami2023landmark} and LongBench\cite{bai2024longbenchbilingualmultitaskbenchmark}.

\subsection{Main Results}

\subsubsection{Memory Usage and Fine-tuning Time}

Table \ref{training-time-memory} compares the memory usage and fine-tuning time of our method with LongLoRA. Our approach significantly reduces memory overhead and fine-tuning time. For example, during 32K length fine-tuning, our method could use stage2+offload optimizer for fine-tuning, while LongLora would OOM. When training with a sequence length of 16K, our memory usage is only 33.5GB, which is significantly less compared to the memory requirements of LongLoRA using 8 GPUs. Moreover, our method achieves comparable or even superior performance with significantly less training time.
\begin{table}
\caption{\textbf{Training Time (8000 steps) and Memory Usage for Different Sequence Lengths.} m128: merge 128 regions of Q. s512: select top512 regions of KV.}
\label{training-time-memory}
\centering
\begin{tabular}{ccccccc}
\toprule
\multicolumn{1}{c}{\makecell{Training \\ Setting}} & \multicolumn{2}{c}{16K (stage2)} & \multicolumn{2}{c}{32K (offload optimizer)} & \multicolumn{2}{c}{50K (stage3)} \\
\cmidrule(r){2-3} \cmidrule(r){4-5} \cmidrule(r){6-7}
& \multicolumn{1}{c}{\makecell{Train \\ Hours}} & \multicolumn{1}{c}{\makecell{Memory \\ (GB)}} & \multicolumn{1}{c}{\makecell{Train \\ Hours}} & \multicolumn{1}{c}{\makecell{Memory \\ (GB)}} & \multicolumn{1}{c}{\makecell{Train \\ Hours}} & \multicolumn{1}{c}{\makecell{Memory \\ (GB)}} \\
\midrule
LoRA & 14.0 & 34.7 & - & OOM & - & OOM \\
LongLora & 11.3 & 34.6 & - & OOM & - & OOM \\
\midrule
\makecell{Ours \\ (m128-s512)} & {11.1} & \textbf{33.5} & 16.1 & \textbf{37.5} & 19.7 & \textbf{51.4} \\
\makecell{Ours \\ (m16-s64)} & \textbf{10.4} & \textbf{33.3} & \textbf{14.7} & {38.0} & \textbf{49.9} & {39.2} \\
\bottomrule
\end{tabular}
\end{table}

\subsubsection{Length Extension: Pretraining}

Tables \ref{tab:ppl_pg19} and \ref{tab:imagenet} show the results of pretraining with MS Attention on PG19 and ImageNet, where our method outperforms various baselines. For PG19, test data is truncated when it is larger than the length of the test.

\begin{table*}[h]
    \centering
    \begin{minipage}{0.48\textwidth}
        \centering
        \caption{Perplexity (PPL) results on the PG19.}
        \label{tab:ppl_pg19}
        \resizebox{\textwidth}{!}
        {\begin{tabular}{lcc}
            \toprule
            Model & Params & PPL (PG19) \\
            \midrule
            TransformerXL & - & 36.3 \\
            Routing Transformer & - & 33.3 \\
            Landmark Attention-200M & 200M & 14.55 \\
            Selection-Merging Attention-200M & 200M & 10.89 \\
            \bottomrule
        \end{tabular}}
    \end{minipage}%
    \hfill
    \begin{minipage}{0.48\textwidth}
        \centering
        \caption{Top-1 accuracy results on ImageNet.}
        \label{tab:imagenet}
        \resizebox{\textwidth}{!}
        {\begin{tabular}{lcc}
            \toprule
            Model & \makecell[c]{FLOPs (G)} & Top-1 Accuracy (\%) \\
            \midrule
            VVT-T-12.9M & 2.0 & 79.4 \\
            Swin Transformer-29M & 4.2 & 81.3 \\
            Biformer-13.1M & 2.2 & 81.4 \\
            Selection-Merging Attention-13.1M & 2.1 & 82.1 \\
            \bottomrule
        \end{tabular}}
    \end{minipage}
\end{table*}

We further conducted experiments on automatic length extrapolation. After training on sequences of 8K and 16K tokens, our method shows stable perplexity even when extending inference to 8x the training length. Table \ref{tab:length_extrapolation} presents the results.

\begin{table}[h]
    \centering
    \caption{Perplexity across different sequence lengths for Select-Merge Attention. Test data is truncated when it is larger than the length of the test.}
    \label{tab:length_extrapolation}
    \begin{tabular}{lccccc}
        \toprule
        Method & 8K & 16K & 32K & 64K & 128K \\
        \midrule
        Select-Merge Attention (8K) & 10.9 & 10.57 & 12.57 & 14.16 & - \\
        Select-Merge Attention (16K) & 9.82 & 9.56 & 10.93 & 12.16 & 16.28 \\
        \bottomrule
    \end{tabular}
\end{table}

\subsubsection{Length Extension: Fine-tuning}

Theoretically, our method augments the positional data through data augmentation, leveraging the relatively strong fitting ability of our algorithm to adapt parameters to arbitrary ratios of NTK-based positional encoding. This allows for the extension to arbitrary positional encoding lengths.

First, for the architecture, during fine-tuning, we replace the attention mechanisms in Llama and Mistral with our MS Attention. For parameter fine-tuning, since only the \(Q\) and \(K\) tokens incorporate positional information, we use LoRA to fine-tune the mapping weights of \(W_Q\) and \(W_K\), while other parameters are only fine-tuned for embeddings and normalization layers. Regarding the choice of positional encoding, we apply a dynamically growing NTK method, starting with an initial scaling factor (set to \(\text{scale}=4096\) in this work), and increase this factor twofold after fine-tuning a small fixed data volume (32M tokens in our case).

For the passkey task, as shown in Table \ref{Passkey task}, using standard NTK positional encoding combined with our algorithm enables length extensions of up to 16x. Moreover, employing random cropping and circular modular sampling stabilizes this extension. Finally, by incorporating dynamic growth, we address the task of passkey generation for arbitrary lengths. For every 4x increase in base length, our approach can solve passkey tasks with a 2x length extension, potentially due to the square root relationship between positional encoding and the base length.

\FloatBarrier
\begin{table}[htbp]
\centering
\caption{Passkey task evaluation: MS Attention with NTK and PI across different token lengths.}
\label{Passkey task}
\resizebox{\textwidth}{!}{
\begin{tabular}{lccccccccc}
\toprule
\textbf{Model} & \textbf{2K-16K} & \textbf{32K} & \textbf{64K} & \textbf{128K} & \textbf{256K} & \textbf{512K} & \textbf{1024K} & \textbf{2048K} & \textbf{4096K} \\
\midrule
\makecell[l]{Llama2-7B-LongLora (ft=100K, PI)} & 1.0 & 1.0 & 1.0 & 0 & 0 & 0 & 0 & 0 & 0 \\
\makecell[l]{Llama2-7B-LongRoPE (ft=256K, PI)} & 1.0 & 1.0 & 1.0 & 1.0 & 1.0 & 1.0 & 1.0 & 0.6 & 0.0 \\
\toprule
\makecell[l]{Llama2-7B-MS (ft=16K, PI, scale 16)} & 1.0 & 1.0 & 1.0 & 0 & 0 & 0 & 0 & 0 & 0 \\
\makecell[l]{Llama2-7B-MS (ft=16K, NTK, scale 4096)} & 1.0 & 1.0 & 1.0 & 1.0 & 0.7 & 0 & 0 & 0 & 0 \\
\makecell[l]{Llama2-7B-MS (ft=16K, CR NTK, scale 4096)} & 1.0 & 1.0 & 1.0 & 1.0 & 1.0 & 0 & 0 & 0 & 0 \\
\makecell[l]{Llama2-7B-MS (ft=16K, CR NTK, scale 16384)} & 1.0 & 1.0 & 1.0 & 1.0 & 1.0 & 1.0 & 0 & 0 & 0 \\
\makecell[l]{Llama2-7B-MS (ft=16K, CRG NTK, scale 4096x2)} & 1.0 & 1.0 & 1.0 & 1.0 & 1.0 & 1.0 & 1.0 & 1.0 & 1.0 \\
\makecell[l]{Mistral-7B-MS (ft=16K, CRG NTK, scale 4096x2)} & 1.0 & 1.0 & 1.0 & 1.0 & 1.0 & 1.0 & 1.0 & 1.0 & 1.0 \\
\bottomrule
\end{tabular}}
\end{table}

A potential issue with this approach is that the positional encoding ratio during inference remains fixed. For instance, when training with a length of 16K, the NTK positional encoding starts with \(\text{scale}=4096\) and the maximum \(\text{position\_ids}=32K\). During inference across all passkey task lengths (0-256K), we use a fixed base scaling rate of 1024 for prediction, achieving 100\% performance. After extending by 16x, the scaling factor becomes \(\text{scale}=4096 \times 16\), and inference using a fixed base scaling rate of 4096 can solve passkey tasks up to 1M in length. Similarly, extending by the corresponding multiples and using a fixed base scaling rate of 65536 extends passkey tasks up to 4M.

However, the use of fixed base scaling rate poses some challenges for perplexity evaluation. In text perplexity testing, we observe that dynamic positional encoding yields better performance as the length varies.

For the long-text perplexity experiments, we evaluated our model on the PG19 datasets, as presented in Table \ref{tab:PG19}. Based on the experimental results, positional interpolation generally yields better performance, as it effectively preserves the learned distributions. In contrast, the NTK-based method may disrupt this distribution to some extent, leading to a performance loss, but it enables much longer extrapolation.

\FloatBarrier
\begin{table}[htbp]
\centering
\caption{Perplexity results on PG19 dataset across different token lengths.}
\label{tab:PG19}
\resizebox{\textwidth}{!}{
\begin{tabular}{lccccccccc}
\toprule
\textbf{Model} & \textbf{16K} & \textbf{32K} & \textbf{64K} & \textbf{128K} & \textbf{256K} & \textbf{512K} & \textbf{1024K} & \textbf{2048K} & \textbf{4096K} \\
\midrule
\makecell[l]{Llama2-7B-LongLora (ft=32K, PI)} &7.35 & 7.22 & - & - & - & - & - & - & - \\
\makecell[l]{Llama2-7B-LongRoPE (ft=256K)} & 7.37 & - & 6.64 &  6.31 & - & - & - & - & - \\
\midrule
\makecell[l]{Llama2-7B-MS (ft=16K, PI, scale 16)} & 6.50 & 6.25 & 6.53 & - & - & - & - & - & - \\
\makecell[l]{Llama2-7B-MS (ft=16K, CR NTK, scale 16384)} & 7.86 & 7.92 & 7.97 & 8.23 & 8.68 & - & - & - & - \\
\makecell[l]{Llama2-7B-MS (ft=16K, CRG NTK, scale 4096x2)} & 7.20 & 7.25 & 7.29 & 7.43 & 8.36 & 9.72 & 14.98 & 26.82 & 57.2 \\
\makecell[l]{Mistral-7B-MS (ft=16K, CRG NTK, scale 4096x2)} & 8.20 & 8.15 & 8.29 & 8.56 & 9.76 & 12.12 & 18.96 & 44.82 & 80.2 \\
\bottomrule
\end{tabular}}
\end{table}

The results show that our model can effectively handle long text sequences, maintaining stable perplexity scores for extended input lengths.

\subsubsection{LongBench}

Table \ref{tab:LongBench} shows Few-shot Learning and Code Completion Evaluation on LongBench: These tasks do not require chat instruct fine-tuning. We have observed significant improvements in "trec", "triviaqa", "lcc", and "repobench-p" using our method on Llama2-7B-4k comparing with Llama2-7B-chat-4k, especially outperforming other models in the LCC task.

\begin{table}[ht]
\centering
\caption{Few-shot Learning and Code Completion Evaluation on LongBench}
\label{tab:LongBench}
\resizebox{\textwidth}{!}{
\begin{tabular}{lccccc}
\toprule
\textbf{Model} & \textbf{TREC} & \textbf{TriviaQA} & \textbf{SAMSum} & \textbf{LCC} & \textbf{RepoBench-P} \\
\midrule
GPT-3.5-Turbo-16k & 68 & \textbf{91.4} & \textbf{41.7} & 54.7 & 53.6 \\
LongChat-v1.5-7B-32k & 63.5 & 82.3 & 34.2 & 53 & \textbf{55.3} \\
XGen-7B-8k & 65.5 & 77.8 & 25.3 & 38.6 & 38.6 \\
InternLM-7B-8k & 52 & 77.8 & 21.2 & 44.1 & 28.8 \\
ChatGLM2-6B-32k & 62.5 & 78.7 & 36.3 & 55.6 & 49.9 \\
Vicuna-v1.5-7B-16k & 71.5 & 86.2 & 40.8 & 51 & 43.5 \\
ChatGLM3-6B-32k & \textbf{79} & 87.1 & 38.2 & 57.66 & 54.76 \\
\textbf{Llama2-7B-chat-4k} & 61.5 & 77.8 & 40.7 & 52.4 & 43.8 \\
\textbf{Llama-7B-32k-MS-PI} & 72.6 & 85.7 & 40.6 & {61.95} & 49.09 \\
\textbf{Llama-7B-16k-MS-CRNTK} & 72.6 & 87.5 & 39.3 & \textbf{65.24} & 54.76 \\
\textbf{Llama-7B-16k-MS-CRGNTK} & 68.6 & 86.2 & 36.6 & {64.1} & 54.3 \\
\textbf{Mistral-7B-16k-MS-CRNTK} & 65.6 & 88.3 & 38.3 & {64.4} & 54.1 \\
\bottomrule
\end{tabular}}
\end{table}

\subsubsection{A controllable convergence for Llama2-7B}

The proposed method allows for controllable convergence and fitting degrees by flexibly setting parameters such as the attention range, the number of selected tokens, and the types of multi-scale compressions, as illustrated in Tabel \ref{tab:loss_comparison}. This adaptability ensures that the model can be fine-tuned to achieve optimal performance across a variety of scenarios, providing robustness and efficiency in handling extended sequences.

\begin{table}[H]
\begin{minipage}{0.46\textwidth}
    \centering
    \caption{Loss for Llama2-7B Using Restricted Scope Multi-Scale Selected Attention.}
    \label{tab:loss_comparison}
    \resizebox{\textwidth}{!}
    {\begin{tabular}{lccccc}
    \toprule
    \textbf{Parameters} & \textbf{loss} \\
    \midrule
    \textbf{MS-patch16-select128-merge16-ntk16} & 1.73 \\
    \textbf{MS-patch16-select32-merge4-ntk16} & 1.71 \\
    \textbf{MS-patch16-select512-merge64-ntk1024} & 1.69 \\
    \textbf{MS-patch8-select16-merge4-ntk1024} & 1.65 \\
    \textbf{MS-patch16-select16-merge1-ntk1024} & 1.63 \\
    \textbf{MS-patch8-select32-merge8-ntk1024} & 1.55 \\
    \textbf{MS-patch8-select32-merge4-ntk1024} & 1.51 \\
    \bottomrule
    \end{tabular}}
\end{minipage}
\begin{minipage}{0.48\textwidth}
    \caption{Ablation experiment for PEFT. PEFT steps: 1000, Evaluation DataSet: PG19 validation. }
    \label{Ablation-peft}
    \centering
    \resizebox{\textwidth}{!}
    {\begin{tabular}{cccccccc}
    \toprule
    \multicolumn{1}{c}{PEFT} & \multicolumn{1}{c}{\makecell{Training Context Length\\ w/wo setting}} & \multicolumn{3}{c}{Evaluation Context Length} \\
    \cmidrule(r){3-5} &  & 32768 & 16384 & 8192 \\
    \midrule
    qkvo & 16384-m16-s64 & 8.96 & 8.33 & 8.32\\
    \midrule
    ko & 16384-m16-s64 & 8.55 & 8.36 & 8.43 \\
    \bottomrule
    \end{tabular}}
\end{minipage}
\end{table}
\subsubsection{ablation experiment}

Firstly, in our pursuit of enhancing efficiency in micro-adjustments, we conducted fine-tuning on \(W_q\), \(W_k\), \(W_v\), and \(W_o\). The resulting Perplexity (PPL) is nearly identical to when fine-tuning is solely applied to \(W_k\) and \(W_o\). In some instances, the latter even yields slightly higher results, as show in Table \ref{Ablation-peft}. This observation suggests that focusing on refining \(W_k\) and \(W_o\) alone can be as effective as fine-tuning all parameters mentioned earlier.

After conducting ablation experiments, this paper demonstrates that full attention can also achieve high extrapolation ratios using a high proportion of NTK encoding, and even the high-ratio PI method can achieve substantial extrapolation. The advantage of our attention mechanism lies in its ability to fine-tune on long sequences, such as 32K tokens. Specifically, during fine-tuning, we employ a subset of KV, which allows us to use less GPU memory and achieve faster processing speeds while maintaining performance comparable to full Attention. This reduces memory consumption and speeds up the process, while still achieving results comparable to full attention. Furthermore, our experiments show that when testing with untrained proportions, our method generalizes better to different ratios compared to full attention. This allows our approach to dynamically adjust the positional encoding growth, giving it a significant advantage in terms of scalability.
\FloatBarrier 

\begin{table}[ht]
\begin{minipage}{0.46\textwidth}
    \centering
    \caption{PPL on the PG19 with Lora or not.}
    \label{PPL Lora or not}
    \resizebox{\textwidth}{!}
    {\begin{tabular}{cccccc}
    \hline
    \textbf{Model Configuration} & \textbf{2K} & \textbf{4K} & \textbf{8K} & \textbf{16K} & \textbf{32K} \\
    \hline
    \text{Llama-7B-MS-no\_LoRA} & 7.06 & 6.96 & 6.78 & 6.56 & 6.93 \\
    \text{Llama-7B-MS-LoRA} & 7.61 & 7.42 & 7.17 & 7.03 & 7.06 \\
    \hline
    \end{tabular}}
\end{minipage}
\begin{minipage}{0.48\textwidth}
    \centering
    \caption{Memory and Time w/wo DeepSpeed and LoRA}
    \label{Memory and Time Lora or not}
    \resizebox{\textwidth}{!}
    {\begin{tabular}{cc}
    \hline
    \textbf{Configuration} & \textbf{Memory (MB) | Time (s/it)} \\
    \hline
    \text{With DeepSpeed \& LoRA} & 36,786 \, | \, 6.37 \\
    \text{Without DeepSpeed} & 31,524 \, | \, 6.19 \\
    \text{Without LoRA} & 35,074 \, | \, 7.28 \\
    \text{Without DeepSpeed \& LoRA} & 40,132 \, | \, 6.38 \\
    \hline
    \end{tabular}}
\end{minipage}
\end{table}

\begin{table}[ht]
\centering
\caption{PPL on the PG19 (50 samples) with MS Attention or not. FT steps: 1000, FT length: 16384, ntk scale: 16384, PI scale: 16.}
\label{PPL MS or not}
\begin{tabular}{cccccc}
\hline
\textbf{Model Configuration} & \textbf{8K} & \textbf{16K} & \textbf{64K} & \textbf{128K} & \textbf{256K} (scale=16384/32768/8192) \\
\hline
\text{Llama-7B-MS-ntk16384} & 7.15 & 9.58 & 9.55 & 22.09 & 55.76/90.92/181.20 \\
\text{Llama-7B-FA-ntk16384} & 7.15 & 9.57 & 9.36 & 21.54 & 53.92/103.03/298.86 \\
\text{Llama-7B-MS-PI16} & 7.01 & 9.23 & 9.18 & - & - \\
\text{Llama-7B-FA-PI16} & 6.99 & 9.18 & 9.04 & - & - \\
\hline
\end{tabular}
\end{table}

\section{Conclusion}

In this paper, we build upon the LongLoRA framework, utilizing a mechanism of selection and merging within the Attention mechanism (referred to as MS Attention). By employing our approach on 2 $\times$ A100 40GB GPUs, we achieve the same level of length extension for Llama2-7B as LongLoRA does on 8 $\times$ A100 80GB GPUs. Specifically, we extend the context length of Llama2-7B to 100K tokens, significantly reducing the resource requirements for handling long sequences. Moreover, through the introduction of recursion, our method maintains stable perplexity even with sequences up to 2M tokens in length. Finally, the flexibility of our MS Attention mechanism allows for adjustable selection size and selection quantity, which, combined with restricting the attention range of each token, enables adaptable fitting and convergence rates.

\newpage

\bibliography{neurips_2024}

\appendix
\newpage
\section{Appendix / supplemental material}
\bf{Analysis of Attention with Correlation-Aware Selection and Merging is presented below:}\label{Analysis}
\subsection{Interpreting Attention from the Perspective of Interpolation}\label{sec:approximate interpolation}
In this section, we reinterpret the theoretical formulation of the Attention mechanism by recasting it as a special interpolation formula. This perspective elucidates the role of different components within Attention and demonstrates its clustering effect. The detailed analysis is as follows:

\subsubsection{Bilinear and Cubic Spline Interpolation}

Consider the following interpolation formula:

\[
f(x,y)=\sum_{i=0}^h\sum_{j=0}^w W_{ij} \cdot f(x+i,y+j)
\]

\[
o(x,y)=\sum_{i=0}^h\sum_{j=0}^w g(d_{ij}) \cdot v(x+i,y+j) \quad \text{or} \quad \sum_{i=0}^h\sum_{j=0}^w g(d_{ij}) \cdot v(x_i,y_j)
\]

In the above equations, \( h = 2 \) corresponds to bilinear interpolation, while \( h = w = 3 \) corresponds to cubic spline interpolation. Here, \( (x,y) \) represents the target interpolation location, and \( d_{ij} \) denotes the distance metric between \( (x_i,y_j) \) and \( (x,y) \). The distance metric \( d_{ij} \) can be computed using various methods, such as the \( n \)-norm or inner product. The function \( g(d_{ij}) \) is a weighting function based on this distance metric. For bilinear interpolation, \( g(d_{ij}) = \frac{x_i-x}{x_i-x_j} \), while for cubic spline interpolation, \( g(d_{ij}) = \frac{x_i-x}{x_i-x_j} \).

By comparing these interpolation formulas with the Attention mechanism, we can approximate the global interpolation formula as follows: Let \( g(d_{ij}) = \text{softmax}(q_{i}k_{j}^T) \) and \( v(x_i,y_j) = v \). This demonstrates that the fundamental principles underlying both interpolation and Attention are similar. Both methods combine vectors with similar features, leading to interpolated results that emphasize these shared features, making them more useful for the task at hand. Although Attention does not strictly satisfy the conditions for interpolation—since it does not require the value at position \( (x_i,y_j) \) to be \( v(x_i,y_j) \)—it can be considered a generalized form of interpolation or regression. This analogy may explain why training in language models is often referred to as autoregression.

\subsubsection{Interpolation Formula on Riemannian Manifolds}

We can further extend this analogy by considering Attention as a special form of interpolation mapping or exponential mapping on a manifold. For instance, on a Riemannian manifold, there exists a matrix exponential mapping given by:

\[
\text{Exp}_{(P, \lambda)}(T) = P^{1/2} \exp{(\lambda P^{-1/2} T P^{-1/2})} P^{1/2}
\]

where \( T \) is a tangent vector matrix, typically a symmetric matrix that can be decomposed into \( ww^T \), and \( P \) is the metric matrix corresponding to a point on the manifold.

By generalizing the variables in the Attention formula, we can map \( ww^T \) to \( w_Q w_K^T \) and the metric matrices \( P^{1/2} \) and \( P^{-1/2} \) to the \( X \) matrices in Attention. Thus, the interpolation formula becomes:

\[
\text{Exp}_{(P, \lambda)}(T) = \exp{(X w_Q w_K^T X)} X^{-1}
\]

This transformation completes the interpolation and maps it to the \( V \) space.

In summary, this theoretical perspective allows us to view the Attention mechanism as a form of interpolation, enabling us to leverage many well-established principles and techniques from interpolation to improve Attention. For example, giving higher weights to points closer in distance is a well-known and effective method in interpolation. This approach is employed in several successful algorithms, such as the Routing Transformer, Landmark Attention, and Biformer. Our MS Attention mechanism, under certain fixed parameter settings, can be nearly equivalent to these methods, fully covering their application scenarios. It can also reduce to other efficient Attention mechanisms, such as Swin Transformer. Detailed analysis of these equivalences is provided in the next section.

\subsection{Approximation and Convergence of Our MS Attention to Various Efficient Attention Methods}\label{Efficient Attention Methods}

Our proposed Select-Merge Attention (MS Attention) can approximate and converge to the optimal solutions of many efficient Attention mechanisms that utilize a subset of key-value (KV) pairs. Below, we describe the computation process using the notations introduced in the algorithm description:

- \bf{Selection:}
\[
A_s = Q_s'K_s'^T, \quad \text{Idx} = \text{topkIndex}(A_s)
\]
- \bf{Merge:} 
\[
Q_s = \text{merge}_q(Q_s'), \quad \text{Idx} = \text{filter}(\text{merge}_q(\text{Idx})), \quad KV_s = \text{Select}(KV, \text{Idx})
\]
- \bf{Final Attention:} 
\[
O = \text{Attention}(Q_s, K_s, V_s)
\]

\subsubsection{Landmark Attention}

Landmark Attention adjusts the attention scores by incorporating landmark tokens, with the output expressed as:
\[
O = (\text{softmax}(QK^T) \cdot \text{repeat}(\text{softmax}(QG^T), \text{blocksize}, \text{dim}=-1))V
\]

When setting the split size of \(Q\) to 1 and the split size of \(K\) to blocksize, our MS Attention also calculates \(\text{softmax}(QK_s^T)\) and \(\text{softmax}(QK_s'^T)\), which fully enables the adjustment of attention scores using semantic tokens. This operation can be approximated under other settings as well.

Alternatively, if the score adjustment is performed without directly multiplying \(\text{softmax}(QK_s'^T)\), treating semantic tokens as regular KV tokens during Attention computation, the output can be expressed as:
\[
\frac{\text{sumexp}(QK_s^T) \times \text{softmax}(QK_s^T)V_s}{\text{sumexp}(QK_s^T) + \text{sumexp}(QK_s'^T)} + \frac{\text{sumexp}(QK_s'^T) \times \text{softmax}(QK_s'^T)V_s'}{\text{sumexp}(QK_s^T) + \text{sumexp}(QK_s'^T)}
\]
where \(V_s'\) is a linear combination of \(V_s\) or \(V\). This approach can also directly adjust the attention coefficient of \(V_s\) in the output, allowing convergence to the optimal solution based on the task.

\subsubsection{BiFormer}

The BiFormer computation process is as follows:
\[
A_r = Q_r(K_r)^T, \quad I_r = \text{topkIndex}(A_r), \quad KV_g = \text{Select}(KV, I_r)
\]
\[
O = \text{Attention}(Q, K_g, V_g)
\]

When the Merge size is set to 1, our method fully degenerates to BiFormer. To achieve lower complexity, BiFormer requires setting the initial region \(Q_r, K_r\) relatively large and then compressing it as a region representative. This approach reduces the accuracy of selection.

When merge size is greater than 1, our method performs more fine-grained partitioning and selection. If BiFormer’s selection is optimal, our algorithm can also converge to this optimal selection. Overall, compared to BiFormer, our method has a larger convergence space, and under the same fine partitioning and selection, our method has relatively lower computational complexity.

\subsubsection{Routing Attention Methods}

For the Routing Transformer, the following update rule is applied:
\[
\mu \leftarrow \lambda \mu + \frac{(1 - \lambda)}{2} \sum_{i:\mu(Q_i)=\mu} Q_i + \frac{(1 - \lambda)}{2} \sum_{j:\mu(K_j)=\mu} K_j
\]
This can be rewritten as:
\[
\mu \leftarrow \lambda \mu + \frac{(1 - \lambda)}{2} \text{argmax}_{qu}(\mu Q^T)Q + \frac{(1 - \lambda)}{2} \text{argmax}_{qu}(\mu K^T)K
\]
where
\[
\text{argmax}_{qu}(\cdot)=
\begin{cases}
1 & \text{argmax}(\cdot, \text{dim}=-2) \\
0 & \text{otherwise} \\
\end{cases}
\]
\[
\text{Idx}_Q = \text{topkIndex}(\mu Q^T), \quad \text{Idx}_K = \text{topkIndex}(\mu K^T)
\]

Using the triangle inequality of the metric, the above approximation can select \(Q\)-related KV tokens with:
\[
\text{Idx} = \text{topkIndex}(Q \mu^T \mu K^T)
\]
Our semantic tokens in each Attention step also cluster similarly:
\[
m_s = \text{Softmax}(m_s K_s^T) K_s
\]
or
\[
m_s = m_s + \text{softmax}(Q_s K_s^T) V_s
\]

Our selection process can be written as:
\[
\text{Idx} = \text{topkIndex}(m_s W_Q W_K^T m_s^T)
\]
where \(m_s\) is the regional semantic token of \(X_s\) (e.g., average or \(m_s = \text{Softmax}(m_s K_s^T) K_s\)), equivalent to further clustering.

Therefore, during selection, the cluster center can be used to represent the tokens within the relevant cluster, similar to the Routing Transformer. Our method, by only using the cluster center for selection, introduces minimal quantization error. However, because the selection quantity is sufficient, the loss is negligible. In contrast, the Routing Transformer loses some related \(Q\) and \(K\) tokens to ensure regular shape, introducing non-negligible error.

\subsubsection{Swin Transformer}

The Swin Transformer utilizes local Attention and shifted local Attention. Our selection mechanism can completely converge to Swin Transformer when it is optimal.

- Local Attention:
\[
\text{softmax}(Q_{i:j}K_{i:j}^T)V_{i:j}
\]
When this is the optimal solution, our selection mechanism will automatically select tokens within the local region.

- Shifted Local Attention:
\[
\text{softmax}(Q_{i:j}K_{i+r:j+r}^T)V_{i+r:j+r}
\]
When this is optimal, our selection mechanism will automatically select tokens within the corresponding region (\(KV_{i+r:j+r}\)), where \(r\) is the cyclic shift offset.

In a similar manner, our method can approximate or cover many variants of the above Transformers.

To compare our method with these efficient Transformers, we conducted experiments on the PG19 datasets, as shown in the tables \ref{Table 4}.

\begin{table}[ht]
\centering
\caption{Perplexity (PPL) and Training Memory Consumption (MB) on PG19. The 20186/22496 means using flashatten or not. The $1 \times 8192$ means $bsz \times seqlen$.}
\label{Table 4}
\begin{tabular}{lccccr}
\toprule
\textbf{Model} & \textbf{PPL (PG19)} & \textbf{1 × 8192} & \textbf{1 × 4096} & \textbf{1 × 2048} \\
\midrule
\textbf{MS Attention-60M} & 13.9 & - & - & - \\
\textbf{MS Attention-200M} & 9.32 & 20186/22496 & 13152/14300 & 9976/10346 \\
\textbf{Landmark Attention-200M} & 14.55 & >40960 (OOM) & >40960 (OOM) & 17938 \\
\bottomrule
\end{tabular}
\end{table}

On the PG19 dataset, our method outperforms Landmark Attention in terms of both performance and efficiency, improving PPL by 4. The efficiency advantage becomes more pronounced as the input length increases.


\begin{table}[htbp]
\centering
\caption{Perplexity results on Proof-Pile dataset across different token lengths.}
\label{tab:Proof-Pile}
\resizebox{\textwidth}{!}{
\begin{tabular}{lccccccccc}
\toprule
\textbf{Model} & \textbf{16K} & \textbf{32K} & \textbf{64K} & \textbf{128K} & \textbf{256K} & \textbf{512K} & \textbf{1024K} & \textbf{2048K} & \textbf{4096K} \\
\midrule
\makecell[l]{Llama2-7B-MS (ft=16K, PI, scale 16)} & - & - & - & - & - & - & - & - & - \\
\makecell[l]{Llama2-7B-MS (ft=16K, CR NTK, scale 4096)} & - & - & - & - & - & - & - & - & - \\
\makecell[l]{Llama2-7B-MS (ft=16K, CRG NTK, scale 4096x2)} & 3.20 & 3.05 & 2.96 & 3.43 & 4.36 & 5.12 & 7.98 & 10.82 & 27.2 \\
\makecell[l]{Mistral-7B-MS (ft=16K, CRG NTK, scale 4096x2)} & - & - & - & - & - & - & - & - & - \\
\bottomrule
\end{tabular}}
\end{table}

\subsection{Detailed Parameter Settings}\label{Detailed Parameter Settings}

Our algorithm can encompass the majority of Q-attention to KV-subset methods by adjusting parameters such as the QKV segment size, the number of selected top-K high-relevance KV regions, and the number of merged Q regions. Below is a detailed discussion on the selection of these hyperparameters:

\subsubsection{QKV Segment Size Selection:}
   The QKV segment size is crucial and typically ranges from 8 to 128. This parameter must be chosen by balancing computational complexity and performance. In future work, we plan to incorporate Triton operators in the QK routing step to achieve linear spatial complexity, thereby mitigating the current limitations:
   
   - (1) During the selection and routing process, a semantic token represents each region, and the relevance between Q and K semantic tokens is measured. KV tokens related to the Q region are then selected based on this relevance. A larger QK region size results in a coarser semantic representation, leading to less precise KV token selection. To minimize this loss, more KV tokens must be selected in the second step, thereby reducing information loss due to coarse semantics. Thus, smaller segment sizes are preferred, though incorporating Triton operators in the QK routing step is anticipated to provide linear spatial complexity.
   
   - (2) Alternatively, a larger segment size can be set, and more KV tokens can be selected in the second step. This approach may increase algorithmic complexity as more KV tokens are likely to be used for interpolation within the same Q region. However, this issue can be alleviated through the third step of merging, where the selected KV tokens can be further merged and selected according to relevance, filtering out irrelevant information.
   
   - (3) Additionally, the segment size of Q regions is independent of the segment size of KV regions. A relationship between them should only be established when specific tasks and complexity constraints require it.

\subsubsection{Top-K High-Relevance KV Region Selection:}
   The number of selected top-K high-relevance KV regions is generally determined by factors such as the model’s pre-training length and the amount of task-specific data. It may also need to be adjusted based on the granularity of the segmentation from the first step. Current large model training methods suggest that selecting 1K-8K KV tokens for interpolation is robust:
   
   - (1) The above represents a general approach, while scenario-specific selection yields higher performance and efficiency. For example, if the task’s data volume is small, a relatively small number of high-relevance regions can be selected to maximize overfitting and memorize the critical data. Conversely, as the data volume increases, more KV tokens may be required for autoregressive prediction of the next token, necessitating the selection of more high-relevance regions.
   
   - (2) In addition to task-based selection, the granularity of segmentation from the first step must also be considered. If segmentation precision is insufficient, more KV tokens should be selected to prevent the loss of critical information. Subsequent merging can then be used to jointly select high-relevance information, improving space and time complexity by sharing KV tokens across Q regions.
   
   - (3) For length fine-tuning, the model's parameters have adapted to the interpolation degree of the pre-training length, which uses a specific number of KV tokens. Hence, the same magnitude of KV tokens must be selected during fine-tuning to avoid overfitting.
   
   This parameter is highly flexible and can be adapted to various scenarios, often requiring some experience for optimal settings. Future work may involve incorporating a loss function to control the selection space, allowing the algorithm to automatically select the appropriate size based on the loss.

\subsubsection{Merging Q Regions:}
   The number of merged Q regions generally depends on the segmentation precision and space complexity. In scenarios where space complexity is not a major concern, sizes ranging from \( \frac{topk}{4C_Q} \) to \( \frac{topk}{C_Q} \) can be chosen. The number of merged KV tokens is typically selected from a range of topk to 2topk KV tokens. Generally, sharing high-relevance KV regions across multiple Q regions results in lower space complexity. Moreover, during the merging process, high-relevance KV regions are selected with flexibility, enhancing performance through methods such as:
   
   - a. Simple unique and sort operations for selection.
   
   - b. Unique operations followed by a secondary segmentation based on high-relevance scores, with allocation and selection according to region clustering.

\subsubsection{Algorithm Complexity Analysis}

(1) Let the segmentation sizes for the Q and KV regions be \( C_Q \) and \( C_K \), respectively. The time and space complexity of routing Q and K using dot products is \( O\left(\frac{N^2}{C_QC_K}\right) \). 

(2) In the second step, the selection of high-relevance regions, denoted by \( C_S \), results in space complexity of \( O\left(\frac{N}{C_Q}C_S\right) \) due to the storage of necessary indices. 

(3) The merging step, with a merge size of \( C_M \), involves combining the selected indices. Filtering algorithms can be introduced during merging; in this work, we employ unique and high-score selection. The space complexity for storing KV tokens after selection and merging is \( O\left(\frac{N^2}{C_QC_M}C_S\right) \). 

(4) The final attention operation has a time complexity of \( O(2\frac{N}{C_QC_M}C_QC_MC_S) = O(NC_S) \). Hence, only the segmentation size and selection quantity affect the algorithm’s complexity.

Without merging, the selected KV tokens have a complexity of \( O\left(\frac{N^2}{C_Q}C_S\right) \), whereas with merging, the complexity is \( O\left(\frac{N^2}{C_QC_M}C_S\right) \). The reduction in complexity due to merging is because the selection process is only carried out after merging, storing only the selected indices before that.

\subsection{Positional Awareness and Breaking Translation Invariance}\label{Positional}

\subsubsection{Positional Awareness:}
Our algorithm leverages semantic token-based routing and selection mechanisms to capture boundary effects, thus disrupting the translation invariance inherent in global Attention mechanisms.

Semantic Token Routing and Selection: 
   The routing process in our algorithm employs semantic tokens, which represent the semantics of a region, carrying contextual information similar to a sliding window in convolutional neural networks (CNNs). Due to the contextual variance surrounding each token, the derived semantic tokens differ, leading to the routing of different KV tokens. This results in the selection of distinct interpolation variables and, consequently, varying outputs. This mechanism allows our approach to effectively capture boundary effects, unlike traditional global Attention which maintains translation invariance.

\subsubsection{Extrapolation through Finetuning and Position Interpolation:}
   Our method can be fine-tuned during training and later applied to full Attention during inference, enabling significantly higher extrapolation factors. Specifically, our MS Attention mechanism can achieve up to twice the extrapolation compared to traditional methods, with potential reasons outlined as follows:

   - \bf{Improved Relative Positional Awareness: }
     The primary reason behind this extrapolation capability is our selection mechanism, which better extends the model's awareness of relative positions. In full Attention, tokens near the boundaries rely on all preceding tokens for regression, potentially leading the model to integrate all previous positional information. In contrast, our selection mechanism integrates a subset of positions into the final positional information, leading to a more accurate representation of relative positions. This method allows the model to assert the correct positional information even with different combinations of preceding positions.

   - \bf{Extrapolation Limits:}
     The reason for successful extrapolation up to twice the training length, but not beyond, lies in the finetuning process itself. During finetuning with a length \( l \), the model learns the correct positional information and its integration within this range. Therefore, the model can correctly extrapolate up to \( 2l \), but further extrapolation may fail as it exceeds the recognized range.

     For instance, when finetuning LLaMA2-7B with a length of \( l = 16K \), the positional range for `Position\_ids` is set from 1 to 16K. By increasing the initial position range, e.g., choosing starting positions within the 1-64K range (\( l_1 = 64K \)), and setting the interpolation ratio for position interpolation as \( 80K/4K = 20 \) or higher, we successfully achieve 100\% accuracy on the 80K-length passkey task. Similarly, by exposing the model to even longer positions and using our adaptive positional interpolation encoding, we set \( l_1 = 128K \) and successfully extrapolate to 144K. This analysis suggests that our method, even with simple position interpolation, has the potential to achieve near-infinite length extrapolation.

\subsubsection{Extrapolation through Finetuning and NTK-Based Positional Encoding:}
Combining our MS Attention with NTK-based positional encoding enables extrapolation by factors exceeding tenfold. This approach is explained using the radix theory proposed by the authors of RoPE:

   - \bf{Learning Relative Magnitudes:}
   Our MS Attention accurately learns the relative magnitudes of positional encodings. By incrementally increasing the radix base size, the model learns to represent the relative magnitudes across different radices. Additionally, by varying the starting positions, the model further refines its understanding of relative magnitudes within the same radix.

   - \bf{Future Work:}
   Future extensions will incorporate relative shifts between Q tokens and K tokens, allowing the model to sense deviations between different positions of Q and K.

This combination of MS Attention and NTK-based encoding showcases a significant potential for enhancing extrapolation capabilities, ultimately pushing the boundaries of positional understanding in large language models.

\subsection{Position Information Learning Process Analysis:}

\subsubsection{Forward Pass}

In the computation process, the steps of Attention typically follow this sequence:

1. First, we introduce two components, \(P_c\) and \(P_s\), to model cosine and sine positional encoding:
\[
K_P = K \odot P_c + K_s \odot P_s
\]
\[
Q_P = Q \odot P_c + Q_s \odot P_s
\]

The matrices \(P_c\) and \(P_s\) are structured as:
\[
P_c = \begin{bmatrix}
\cos \theta_0 & \cos \theta_0 & \dots & \cos \theta_{d/2-1} \\
\cos 2\theta_0 & \cos 2\theta_0 & \dots & \cos 2\theta_{d/2-1} \\
\vdots && \vdots \\
\cos n\theta_0 & \cos n\theta_0 & \dots & \cos n\theta_{d/2-1}
\end{bmatrix}
\]

\[
P_s = \begin{bmatrix}
\sin \theta_0 & \sin \theta_0 & \dots & \sin \theta_{d/2-1} \\
\sin 2\theta_0 & \sin 2\theta_0 & \dots & \sin 2\theta_{d/2-1} \\
\vdots && \vdots \\
\sin n\theta_0 & \sin n\theta_0 & \dots & \sin n\theta_{d/2-1}
\end{bmatrix}
\]

Moreover, we apply a learned matrix \(W_s\) to \(K_s\) as follows:
\[
K_s = K \times W_s = K \times \begin{bmatrix}
\begin{matrix} 0 & -1\\ 1 & 0 \end{matrix} & &0 \\
& \ddots & \\
0 & & \begin{matrix} 0 & -1\\ 1 & 0 \end{matrix}
\end{bmatrix}_{n \times n}
\]

2. Compute similarity scores:
\[
\text{Scores} = \frac{Q_PK_P^\top}{\sqrt{d_k}}
\]
3. Apply Softmax:
\[
\text{Weights} = \text{softmax}(\text{Scores})
\]
4. Compute the output:
\[
\text{Output} = \text{Weights} \times V
\]

\subsubsection{Backward Pass}

During the backward pass, we need to compute the gradients for each intermediate variable. Below are the key steps:

2.1 Gradient with respect to \(V\)

First, we calculate the gradient of the loss with respect to \(V\):
\[
\frac{\partial L}{\partial V} = \text{Weights}^\top \frac{\partial L}{\partial \text{Output}}
\]

2.2 Gradient with respect to the weight matrix \(\text{Weights}\)

Next, we compute the gradient of the loss with respect to the weight matrix:
\[
\frac{\partial L}{\partial \text{Weights}} = \frac{\partial L}{\partial \text{Output}} V^\top
\]

2.3 Gradient with respect to the similarity scores \(\text{Scores}\)

Since the weight matrix is obtained through a Softmax function, we need to compute the gradient of the loss with respect to \(\text{Scores}\). The derivative of the Softmax involves the Jacobian matrix, as given by:

\[
\frac{\partial L}{\partial \text{Scores}_{ij}} = \sum_{k} \frac{\partial L}{\partial \text{Weights}_{ik}} \cdot \text{Weights}_{ik} \cdot (\delta_{jk} - \text{Weights}_{ij})
\]

where \(\delta_{jk}\) is the Kronecker delta function.

2.4 Gradients with respect to \(Q\) and \(K\)

Finally, we compute the gradients with respect to \(Q\) and \(K\):

\[
\frac{\partial L}{\partial Q_P} = \frac{1}{\sqrt{d_k}} \cdot \frac{\partial L}{\partial \text{Scores}} \cdot K_P
\]

\[
\frac{\partial L}{\partial K_P} = \frac{1}{\sqrt{d_k}} \cdot Q_P^\top \cdot \frac{\partial L}{\partial \text{Scores}}
\]

2.5 Learning of Final \(QK\) Weights

Parameter Updates and Functional Operators

We first define the position-aware key parameter update as:

\[
K_P = K \left(1 + \times \left(W_S \odot \frac{P_s}{P_c}\right)\right) \odot P_c
\]

The expression 

\[
\left(1 + \times \left(W_S \odot \frac{P_s}{P_c}\right)\right) \odot P_c
\]

can be abstracted as a functional operator denoted by \(\hat{F}\). In a more general form, the operator \(\hat{F}\) could follow a design similar to that of a multi-layer perceptron (MLP), where the key \(K\) is flattened into a 1D vector for a large sparse matrix transformation. It is then reshaped back into an \(n \times d\) dimensional matrix. The sparse large matrix is a block-diagonal matrix where each block consists of sinusoidal functions from the RoPE mechanism, defined as:

\[
D_{m \times d+i} = \begin{bmatrix}
\cos(m\theta_i) & -\sin(m\theta_i) \\
\sin(m\theta_i) & \cos(m\theta_i)
\end{bmatrix}_{2 \times 2}
\]

Gradient Computation

We compute the gradients as follows:

\[
\frac{\partial L}{\partial K_P} = \frac{1}{\sqrt{d_k}} \cdot Q_P^\top \cdot \frac{\partial L}{\partial \text{Scores}}
\]

For the key \(K\), the gradient is:

\[
\frac{\partial L}{\partial K} = \frac{\partial L}{\partial K_P} \odot P_c + \left(\frac{\partial L}{\partial K_P} \times W_S^\top\right) \odot P_s
\]

The gradient with respect to the weight matrix \(W_K\) is given by:

\[
\frac{\partial L}{\partial W_K} = X^\top \frac{\partial L}{\partial K}
\]

Substituting the expression for \(\frac{\partial L}{\partial K}\):

\[
\frac{\partial L}{\partial W_K} = X^\top \left(\frac{\partial L}{\partial K_P} \odot P_c + \left(\frac{\partial L}{\partial K_P} \times W_S^\top\right) \odot P_s\right)
\]

\[
= X^\top \left(\left(Q_P^\top \times \frac{\partial L}{\partial \text{Scores}}\right) \odot P_c + \left(\left(Q_P^\top \times \frac{\partial L}{\partial \text{Scores}}\right) \times W_S^\top\right) \odot P_s\right)
\]

This can be further simplified as:

\[
= X^\top \left(\left(\left(Q \odot P_c + Q_S \odot P_s\right)^\top \times \frac{\partial L}{\partial \text{Scores}}\right) \odot P_c + \left(\left(\left(Q \odot P_c + Q_S \odot P_s\right)^\top \times \frac{\partial L}{\partial \text{Scores}}\right) \times W_S^\top\right) \odot P_s\right)
\]

\[
= X^\top F^\top Q^\top \left(\frac{\partial L}{\partial \text{Scores}} + \frac{\partial L}{\partial \text{Scores}} \times W_S^\top \odot \frac{P_s}{P_c}\right) \odot P_c
\]

\[
= X^\top F^\top Q^\top F_l'
\]
This final form can be abstracted as a functional operator similar to the one defined above.

\subsubsection{Length Extrapolation:}

Let the original sequence length be \(M\), and consider the attention matrix:

\[
A = Q_M K_M^\top = Q \hat{F}_M \hat{F}_M^\top K^\top
\]

We aim to adjust position information for unseen positions using the learned adjustment \(F_l = \left(1 + W^{-1} W' F_l'\right)\):

\[
A = Q_L K_L^\top = Q F_l \hat{F}_M \hat{F}_M^\top F_l^\top K^\top
\]

From previous observations, we know that the adjustment is able to modulate position shifts up to:

\[
\frac{(M-1)}{base^{\frac{i}{2d}}}
\]

Thus, to generalize to unseen tokens, we propose adjusting the actual token positions so that they fall within this range.

\subsubsection{Strategies for Generalization:}

1. Increase M: Increasing the length of the original sequence \(M\) allows us to cover a broader range of positions.

2. Adjust Token Positions: Using a method like Position Interpolation (PI), we can adjust the token positions to fall within the range that the model has learned.

3. Increase Base: Using methods such as NTK-aware RoPE, we can increase the base value. Since the positional base is inherently discrete across dimensions, it is only necessary to adjust a subset of the positional encodings correctly to achieve proper adjustment. An advantage of this method is the reuse of previously learned positional bases. For example, bases like \(16^{\frac{2}{8}}\) and \(256^{\frac{1}{8}}\) can share learned weights, allowing us to generalize to various sequence lengths dynamically.

4. Dimensional Generalization: Similarly, we can extend the generalization mechanism across sequence dimensions. By selectively using parts of the position encodings that fall within the learned range, we can efficiently cover new positions while maintaining computational efficiency.

5. Scaling Numerator and Denominator: We can also consider scaling both the numerator and denominator of position encodings to achieve a more generalized weighting over positional information.

\subsubsection{Other Extrapolation Error Analysis:}

To analyze the error during extrapolation, we assume a linear relationship between different segments, where each part is responsible for the corresponding position transformation. Specifically, the part of the gradient involving \(P_c\) is responsible for handling the ground-truth \(P_c\). Let the positional difference between seen and unseen sequences be:

\[
Q_Z - Q_L = X W_{QZ} - X \left(W_Q + \frac{\partial L}{\partial W_Q}\right)
\]

Now, consider a fine-tuning setup where we train on sequences of length \(M\). For the first \(M\) tokens, the model performs without loss. However, for tokens beyond \(M\), the extrapolation requires additional adjustments. For the first \(M\) tokens, the gradient is:

\[
X_{:M} W_{QZ} = X_{:M} \left(W_Q + \frac{\partial L}{\partial W_Q}\right) + L(V, K)
\]

For the tokens beyond \(M\), the extrapolation error is:

\[
(X_{M:} W_{QZ}) \odot P_c - (X_{M:} (W_Q + \frac{\partial L}{\partial W_Q})) \odot P_c
\]

Letting it generalize continuously over a fixed length enables the continuous reduction of the above error L, thus reducing the extrapolation of the final position information.

\subsection{MS recursion}\label{MS recursion}

\textbf{Recursive Methods}

To further extend sequence length and even achieve infinite generation, many recent studies have adopted recursive methods. These methods apply attention mechanisms recursively, generating longer sequences by iteratively attending to subsets of tokens. By recursively expanding the attention range, these methods can generate sequences that exceed the model's maximum sequence length, thereby breaking the sequence generation limits in LLMs. Examples include Infini-Attention\cite{munkhdalai2024leave}, Transformer-XL\cite{dai2019transformerxl}, and SSM\cite{smith2023simplified}\cite{gu2022efficiently}\cite{sun2023retentive}\cite{gu2023mamba}, which utilize recursive methods during training and inference to achieve extremely long sequence modeling.

To seamlessly use recursive computation for the entire sequence attention, we set each token to attend to a certain range and expand its view progressively with each layer, thereby achieving effective modeling of long sequences.

We utilize a multi-scale MS attention mechanism where each scale attends to different ranges of key tokens, and even within the same scale, different ranges can be set for the key tokens. Since the attention range is fixed within each layer and all operations before computing Softmax are linear, a compress-then-compute strategy can be employed. Specifically, the previous state is stored recursively across all fixed ranges, compressed according to the scale, ultimately achieving long sequence generation with fixed memory size.

For example, we divide the sequence into chunks of length 16,384, denoted as \( X_i \). We then apply three scales of Attention with compression ratios of \{1, 2, 4\} and attention ranges of \{4096, 8192, 16384\}. Each of these attention ranges is compressed according to its respective ratio, resulting in dimensions \{4096, 4096, 4096\}, represented as \( X_{i}^{p1} \), \( X_{i}^{p2} \), and \( X_{i}^{p4} \). These compressed representations are concatenated to form \( X_{i}^{p} \). If the size of \( X_{i}^{p} \) exceeds the maximum attention range of 16,384, we use the uncompressed \( X_i \) directly. This concatenated or original representation is then used as the hidden state input to the next block. The computations for the next block are as follows:

\[
K_i = X_{i}^{p} W_k, V_i = X_{i}^{p} W_v, Q_{i+1} = X_{i+1} W_q, K_{i+1} = X_{i+1} W_k, V_{i+1} = X_{i+1} W_v,
\]

The output of the next block is then calculated by, $cc_{i}$ is same as before:

\[
O_{i+1} = \frac{cc_{i}}{c_{i+1} + cc_{i}} \text{softmax}(Q_{i+1} K_{i}^T) V_{i} + \frac{c_{i+1}}{c_{i+1} + cc_{i}} \text{softmax\_withmask}(Q_{i+1} K_{i+1}^T) V_{i+1}.
\]

\begin{figure}[H]
  \centering
  \includegraphics[width=0.9\textwidth]{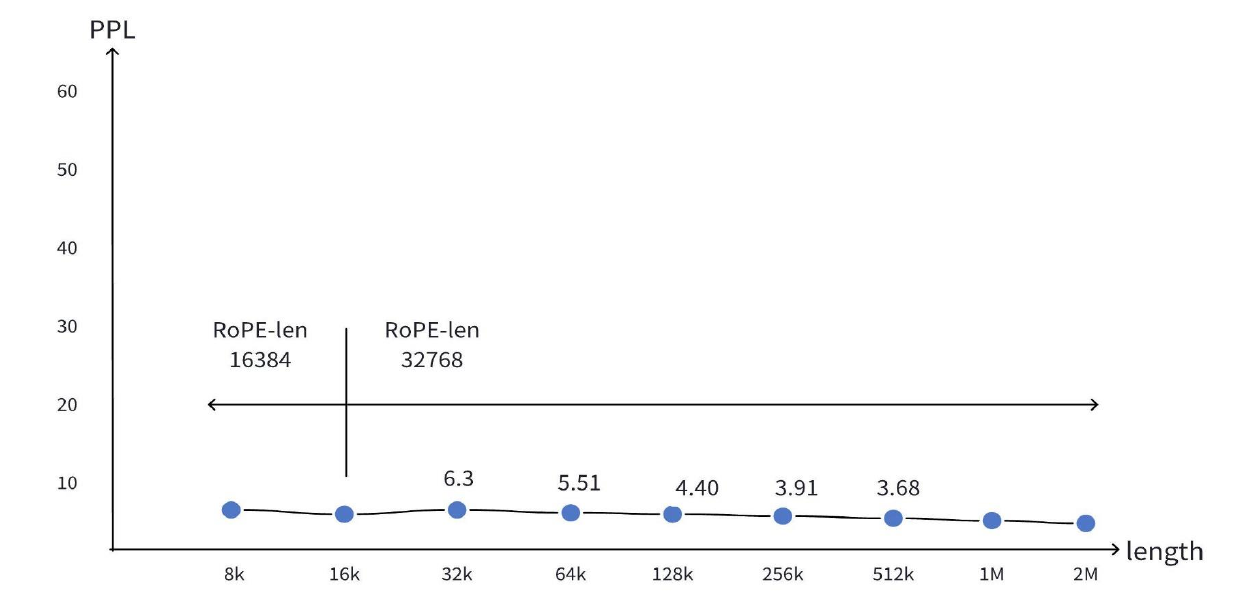}
  \caption{\textbf{Sequence Length Extension Using Recursive Methods.} By combining multi-scale MS Attention with recursive methods, we extend the sequence length. The model is fine-tuned on 16K length sequences, and during evaluation, lengths less than 16K are interpolated to 16K positions, while lengths greater than 16K use 32K position interpolation.}
  \label{Recursive extension}
\end{figure}

\end{document}